\documentclass[runningheads]{llncs}
\usepackage[T1]{fontenc}
\usepackage{caption}
\usepackage{subcaption}
\usepackage{graphicx}
\usepackage{tabularx}
\usepackage[caption=false]{subfig}
\usepackage{svg}
\usepackage{array}        
\usepackage{makecell}
\usepackage{amsmath}
\usepackage{booktabs}
\usepackage{makecell}

\begin{document}

\title{Towards Structured Knowledge: Advancing Triple Extraction from Regional Trade Agreements using Large Language Models}
\titlerunning{Triple Extraction using Large Language Models}
%
\author{Durgesh Nandini\orcidID{0000-0002-9416-8554} \and
Rebekka Koch\orcidID{0009-0002-2174-8352} \and 
Mirco Schönfeld\orcidID{0000-0002-2843-3137}}


%
\authorrunning{Nandini et al.}
%
\institute{University of Bayreuth, Bayreuth, Germany \\
\email{durgesh.nandini@uni-bayreuth.de}\\}
\maketitle              
\begin{abstract}
This study investigates the effectiveness of Large Language Models (LLMs) for the extraction of structured knowledge in the form of Subject-Predicate-Object triples. We apply the setup for the domain of Economics application. The findings can be applied to a wide range of scenarios, including the creation of economic trade knowledge graphs from natural language legal trade agreement texts. As a use case, we apply the model to regional trade agreement texts to extract trade-related information triples. In particular, we explore the zero-shot, one-shot and few-shot prompting techniques, incorporating positive and negative examples, and evaluate their performance based on quantitative and qualitative metrics. Specifically, we used Llama 3.1 model to process the unstructured regional trade agreement texts and extract triples. We discuss key insights, challenges, and potential future directions, emphasizing the significance of language models in economic applications.

\end{abstract}

\section{Introduction}
\label{intro}
\vspace{-2mm}
The evolving landscape of web engineering increasingly demands intelligent systems that can process, structure, and reason over vast, heterogeneous text data published online. Legal, economic, and policy documents are now routinely disseminated through digital platforms, yet remain largely unstructured and inaccessible for automated processing. In response to this challenge, LLMs have emerged as powerful tools for enabling intelligent information extraction at web scale. LLMs have opened avenues for significant techniques of knowledge extraction and processing because of their capabilities to offer pre-trained architectures that have captured vast linguistic and semantic nuances from a wide array of sources. A key technique that enhances their adaptability is prompt engineering \cite{liu2023pre,lester2021power,brown2020language}
which allows models to be directed toward specific tasks without fine-tuning, making them ideal for dynamic, web-based knowledge systems.

Prompt engineering is a pivotal method that extends the capabilities of large language models (LLMs) \cite{sahoo2024systematicsurveypromptengineering} towards crafting task-specific inputs as \textit{prompts} that coax outputs from language models. 
For tasks such as triple generation where subject-predicate-object structures are extracted from unstructured texts, prompt engineering becomes essential because it allows to elicit structured knowledge from LLMs without the need for additional fine-tuning, thus reducing computational resources and expediting experimentation. This approach has become even more valuable in complex domains, where language is ambiguous, and contextual understanding is vital \cite{nandini2024multidimensional}. This study aims to explore the efficiency of language models, in particular the Llama 3.1 \cite{touvron2023llama} model, for subject-predicate-object entity triple extraction from natural language economics legal regional trade agreements. 
Our work specifically tailors the extraction techniques of economic trade-related triples from regional trade agreements (RTAs) that contain highly formalized, structured yet implicit economic obligations, complex multi-party agreements, and domain-specific terminologies that require nuanced interpretation. Standard information extraction models often struggle with these subtleties, making our adaptation of prompt-based extraction particularly valuable. \\
Therefore, the \textbf{main contribution} of our work is the development of large language model prompt pipeline to extract triples from legal regional trade agreement documents. To the best of our knowledge, triple extraction using large language models (LLMs) is a relatively new field within the domain of economic trade exchange transactions. Our approach contributes to the integration of LLM-based extraction pipelines into web-oriented knowledge infrastructures. To implement this, we adopt the zero-shot \cite{brown2020language}
and the few-shot \cite{brown2020language}
prompt engineering methods \cite{brown2020language}
The extracted triples can serve as foundational building blocks for semantic web applications such as linked data generation, legal knowledge graphs, compliance monitoring tools, and intelligent search engines. These applications lie at the intersection of AI and web engineering, emphasizing the growing role of LLMs in enriching and structuring web-based information.  

The rest of the paper is organised as follows: in Section \ref{lit} we briefly discuss the related works and highlight the significance of this study. In Section \ref{methodology}, we describe the dataset used and the methodology that we have used for this study. In section \ref{experiments} we discuss the experimental setup. We present the results and evaluation in section \ref{eval}. At last, in Section \ref{conclusion}, we have the conclusion and the limitations of our work.  

\section{Related Work}
\label{lit}
\vspace{-2mm}
The use of knowledge extraction is a relatively young area of research in the field of Economic trade transactions. However, other related fields such as e-commerce have lightly used knowledge graphs and triples for their studies. The AlimeKG framework \cite{li2020alimekg} is focused on the e-commerce domain where the authors introduce a framework for KG construction in the e-commerce domain by integrating NLP components such as named entity recognition (NER) and relation extraction (RE), facilitating a semi-automated process for knowledge acquisition and validation. Similarly, Yu et al. developed FolkScope \cite{yu2022folkscope}, a framework combining LLMs with human-in-the-loop annotations to build an intention knowledge graph for e-commerce, demonstrating the potential of LLMs in uncovering latent relationships from textual product data. We also identified alternative pipelines, other than LLM based, proposed in other domains. Dessì et al. \cite{dessi2022scicero} explored Transformer models to automatically extract entities from scientific texts and generate a KG. Within the economic and trade domain, Nandini et al. proposed KonecoKG \cite{nandini2024multidimensional}, a multidimensional economic knowledge graph for international trade, highlighting the need for domain-adaptive models and linked data frameworks to capture the complexities of trade agreements. Liu et al. introduced K-BERT \cite{liu2020k}, an early attempt to integrate external knowledge into transformer-based models to improve factual consistency and performance on KG-related tasks. More aligned with prompt-based extraction, Yao et al. proposed KG-BERT \cite{yao2019kg}, which treats triples as textual sequences, applying pre-trained transformers to perform relation classification and triple prediction.

Despite these efforts, we see that there is a gap when it comes to utilising knowledge extraction for econometric trade scenarios. Our study builds upon this emerging intersection by evaluating different prompting methods \cite{iceis25}.

\vspace{-2mm}
\section{Methodology}
\label{methodology}
\begin{figure}
\centering \includegraphics[width=\textwidth, height=3cm]{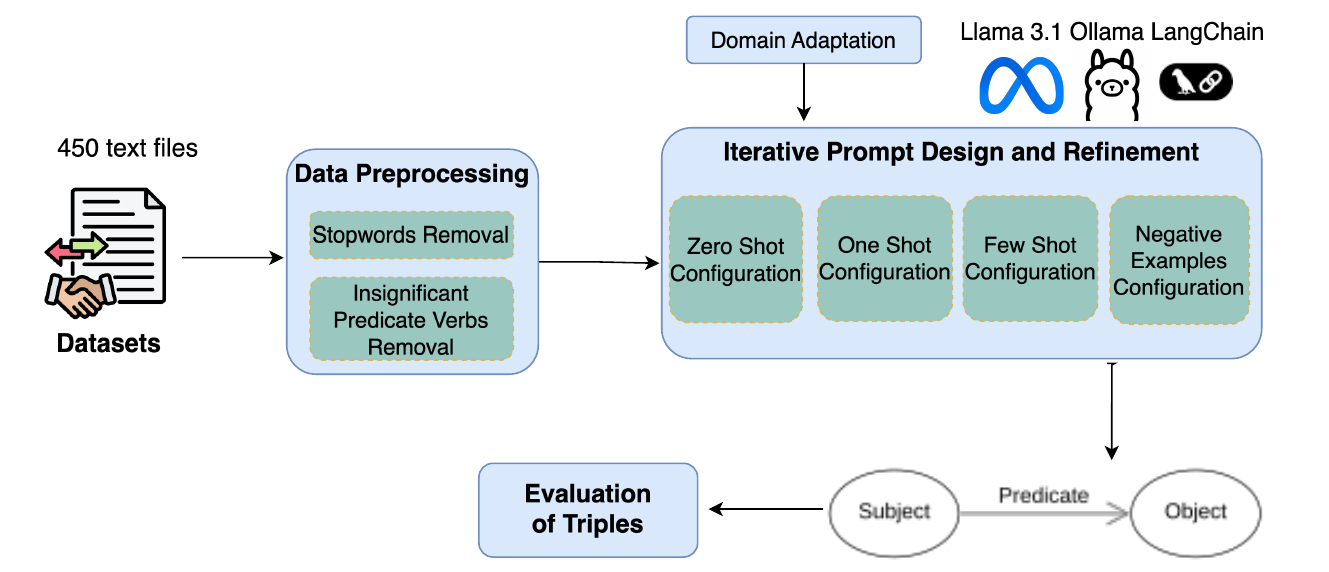}
\caption{Overview of the research methodology} \label{fig:methodology}
\end{figure}

\vspace{-2mm}
In this section we describe the methodology that we have employed for the experimental purposes. To implement and evaluate the zero shot and few shot prompting techniques, we iteratively fine tuned the prompts at different stages. Figure \ref{fig:methodology} summarises the flowchart of the methodology and we define each step of the methodology in the following subsections.  

\noindent \textbf{Dataset and Data Preprocessing} 
We use the regional trade agreement dataset by Alschner et. al \cite{alschner2017text}. Their corpus is based on the  WTO Regional Trade Agreements Information System data containing 450 XML files. Each XML file consists of trade agreement between two countries and is composed of multiple Articles and Chapters. The trade agreements texts are categorised into several sectors, such as agriculture, customs, trade in services, and institutions, etc.
Since the raw data that we have used is a collection of large natural language texts, we cleaned it to increase the efficiency for the model execution. In particular, we removed stopwords and some commonly occurring trade related terms that appear frequently in the dataset but are not semantically significant when extracting triples from the texts. \\
\textbf{Iterative Prompt Design and Refinement} The core of our methodology involved creating and refining four types of prompts, each with increasing task-specific guidance. Starting with a general base prompt, we instructed the model to identify triples within the text. This initial prompt was intentionally broad to gauge the model’s baseline performance in a zero-shot setting. In this prompt, the model received minimal guidance on the structure of triples, aiming to determine its capacity to infer task requirements with limited instruction. Each subsequent prompt was progressively refined to include specific instructions that aligned with the nuances of legal text in trade agreements. For example, we added directives for the model to focus on economic trade-related verbs (e.g., “agree,” “sign,” “ratify,” “export,” “import”) to enhance the relevance of generated triples within the context of trade agreements. Additionally, later prompts emphasized the identification of entities such as contracting parties and economic terms specific to trade, guiding the model to capture legally significant information. By iteratively building upon each prompt, we moved from general language processing toward targeted knowledge extraction that suited the specialized legal domain.

\noindent \textbf{Domain Adaptations and Benchmark Triple Set} Given the specificity of our corpus, legal documents focused on trade agreements, we adapted our prompts to capture the complexity and formality inherent in legal language. To this end, later prompt iterations included instructions to prioritize predicates tied to economic trade-related actions, as these are central to the semantics of trade agreements. Emphasizing such domain-specific verbs helped orient the model toward identifying relationships crucial to understanding trade obligations, rights, and entities within legal texts. 
Subsequently, we created a benchmark dataset of triples generated manually by a domain expert specializing in economics and trade data. The expert manually curated 100 triples, which serve as the ground truth for evaluation. 

\noindent \textbf{Evaluation} The comparison between zero-shot and few-shot prompts enables us to evaluate whether including examples improves the accuracy of extracted triples, particularly in handling domain-specific language. To assess model performance, we employ both quantitative and qualitative evaluation methods. Quantitative metrics, such as precision, recall, F1-score, exact match, partial match, and semantic similarity, provide an objective, reproducible measure of alignment with ground-truth annotations. However, these metrics may not fully capture partial correctness or semantic nuances, especially in complex legal-economic texts. Hence, qualitative evaluation complements this by incorporating human judgment to assess the relevance, interpretability, and domain suitability of the triples. For qualitative evaluation, we propose a curated set of metrics and manually assess 100 randomly sampled triples from each model output. These qualitative metrics are discussed in detail in Section \ref{eval}.

\vspace{-2mm}
\section{Experiments}
\label{experiments}
\vspace{-2mm}
In our experiments, we utilized the LLaMA 3.1 model containing 70 billion parameters. A Python-based program was developed using Ollama \footnote{\url{https://ollama.com/library}} and LangChain \footnote{\url{https://www.langchain.com/}} to execute the language models for triple extractions, and the outputs were subsequently stored for evaluation purposes. For LLM optimization, parameters such as temperatures and prompt strategies play an essential role. 

We have experimented with five types of prompts. We start with a generic prompt. This is the zero shot prompt configuration, wherein we instruct the model to extract subject-predicate-object triples from the texts. We define that the subject and the object must be Named Entities \cite{rau1991,chiu2016ner,nadeau2007survey} while the predicates must be English language verbs. Then, for each subsequent prompt we add further instructions and examples. In the second prompt, we add one example and the definition of Named Entity Recognition (NER). This is the one shot prompt configuration. For the third prompt, we enhance the second prompt by adding a few more examples. This is the few shot configuration. In the fourth prompt, we add a few more examples of what the triples should look like. Alongwith that, we also add a few negative examples and negated instructions. This is the negative examples configuration. With negated instructions intend to add examples that suggest the model what would be a wrong outcome. As a feedback from the results generated by prompt 3, we also instruct the model to not include any verbs that we deem are not significant and to refine the results that we deem might require more information. For example, if the model generates triples such as \textit{'Parties', 'signed','contract'} we instruct the model to define what the term \textit{'Parties'} stand for. We also ask the model to deal with coreference resolution and observe the results obtained. An example of each prompt configuration will be provided in the open code access. 

\vspace{-2mm}

\begin{table}[!htbp]
\centering
\caption{Comparison of Llama 3.1 performance across different prompts}\label{quant}
\begin{tabular}{l @{\hspace{10pt}} c @{\hspace{10pt}} c @{\hspace{10pt}} c @{\hspace{10pt}} c}
\toprule
\textbf{Metric} & 
\makecell{\textbf{Zero Shot} \\ \textbf{Model}} & 
\makecell{\textbf{One Shot} \\ \textbf{Model}} & 
\makecell{\textbf{Few Shot} \\  \textbf{Model}} & 
\makecell{\textbf{Negative Examples} \\ \textbf{Model}} \\
\midrule
\multicolumn{5}{l}{\textit{Exact Match}} \\
\quad Precision     & 0.04 & 0.11 & 0.25 & 0.39 \\
\quad Recall        & 0.22 & 0.38 & 0.57 & 0.66 \\
\quad F1 Score      & 0.07 & 0.17 & 0.35 & 0.49 \\
\multicolumn{5}{l}{\textit{Semantic Match (using embeddings)}} \\
\quad Precision     & 0.06 & 0.14 & 0.30 & 0.46 \\
\quad Recall        & 0.28 & 0.44 & 0.65 & 0.78 \\
\quad F1 Score      & 0.10 & 0.21 & 0.41 & 0.57 \\
\bottomrule
\end{tabular}
\end{table}

\vspace{-2mm}

The output of each prompt execution is a set of subject-predicate-object triples, where the subject and object are named entities identified by the Named Entity Recognition, while the predicates are English language verbs. We have limited the output to 1000 triples per country pair document. We also observe the results from each prompt and use them to create new prompts, implying an indirect feedback to the model. We then create predicate frequency charts, and heatmaps to evaluate the model output and compare them to the benchmark triple set curated by Economics domain experts. We then evaluate the results generated by the model through various metrics.

\section{Results and Evaluation}
\label{eval}
\vspace{-2mm}

\begin{figure}
     \centering
     \begin{subfigure}[b]{0.24\textwidth}
         \centering
         \includegraphics[width=\textwidth, height=1.6cm]{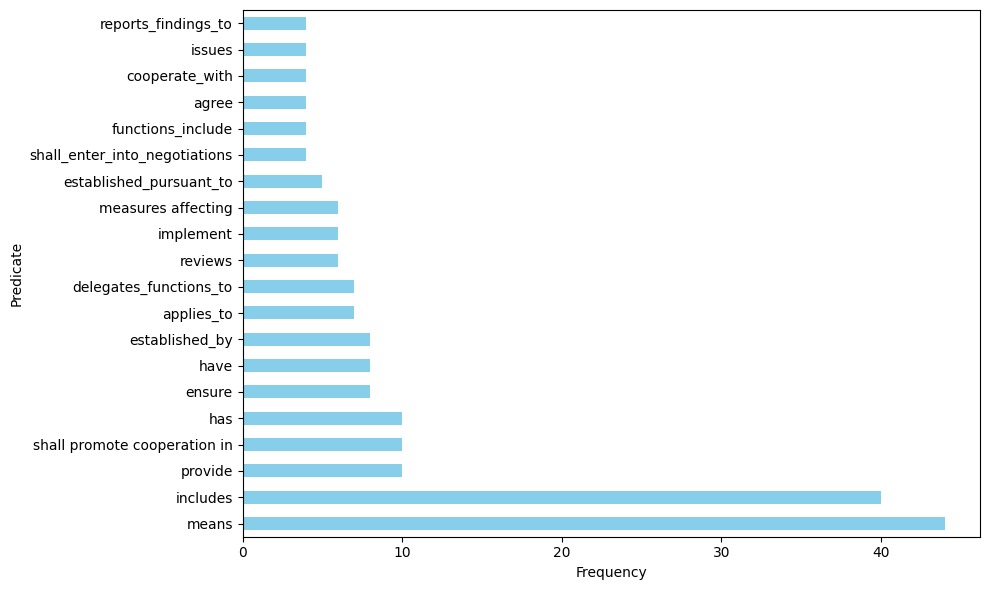}
         \caption{}
         \label{fig:diagram1}
     \end{subfigure}
     \hfill
     \begin{subfigure}[b]{0.24\textwidth}
         \centering
         \includegraphics[width=\textwidth, height=1.6cm]{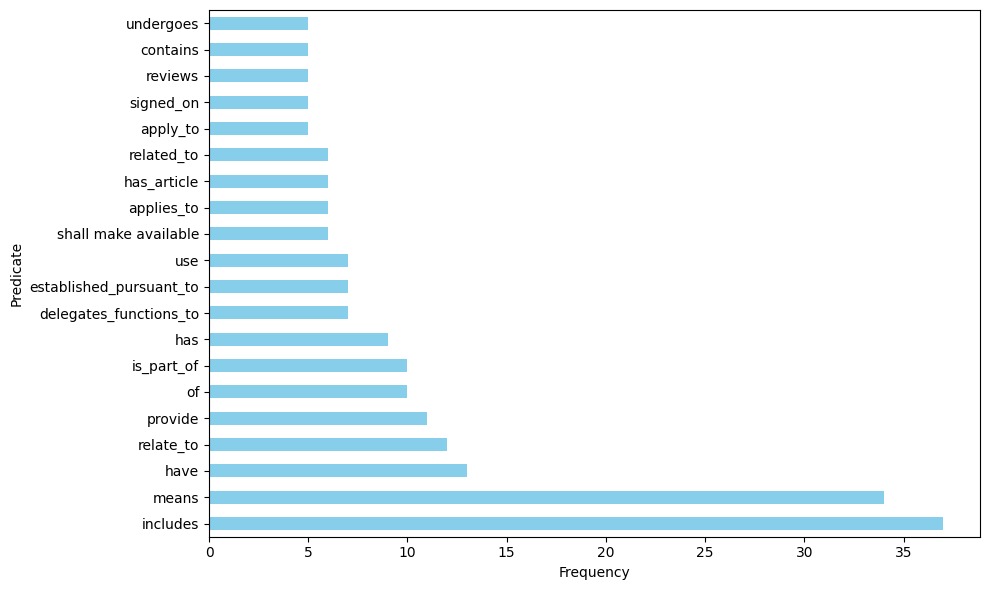}
         \caption{}
         \label{fig:diagram2}
     \end{subfigure}
     \hfill
     \begin{subfigure}[b]{0.24\textwidth}
         \centering
         \includegraphics[width=\textwidth, height=1.7cm]{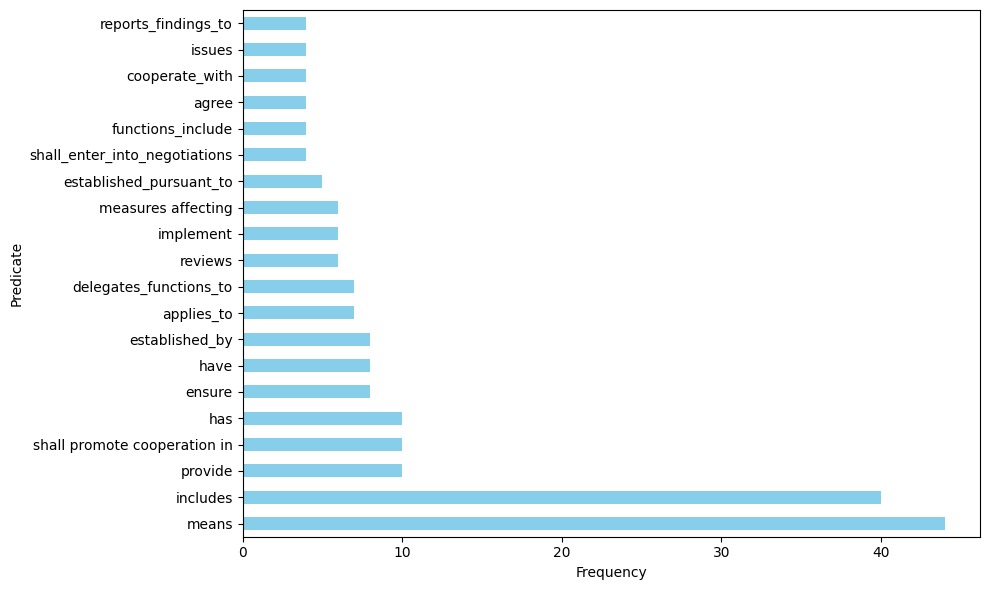}
         \caption{}
        \label{fig:diagram3}
     \end{subfigure}
     \hfill
     \begin{subfigure}[b]{0.24\textwidth}
         \centering
         \includegraphics[width=\textwidth, height=1.7cm]{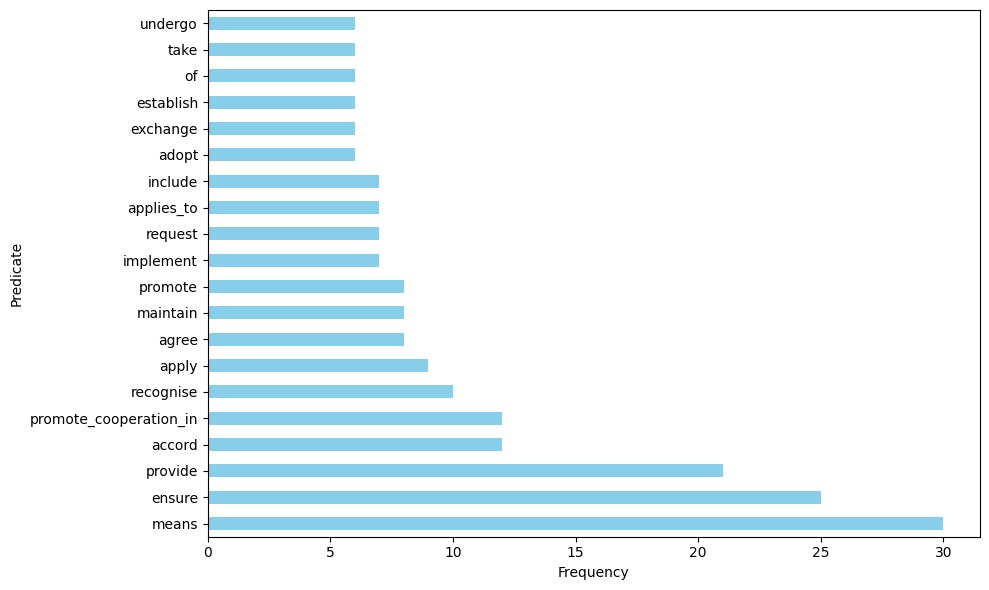}
         \caption{}
        \label{fig:diagram4}
     \end{subfigure}
        \caption{Predicate frequency charts: (a) Zero Shot, (b) One Shot, (c) Few Shot, (d) Negative Examples}
        \label{fig:four_diagrams}
\end{figure}

We evaluate the extracted triples using both quantitative and qualitative metrics. Table \ref{quant} presents the accuracy of four models compared to a domain-expert curated dataset. We also analyze predicate frequency, which informs prompt design by highlighting commonly used and foundational predicates like \textit{means} or \textit{includes}. These insights support refining models to better capture both broad and domain-specific concepts. Figure \ref{fig:four_diagrams} displays bar charts of predicate frequency, illustrating their distribution in economic trade agreements.

\begin{figure}
     \centering
     \begin{subfigure}[b]{0.24\textwidth}
         \centering
         \includegraphics[width=\textwidth, height=2.5cm]{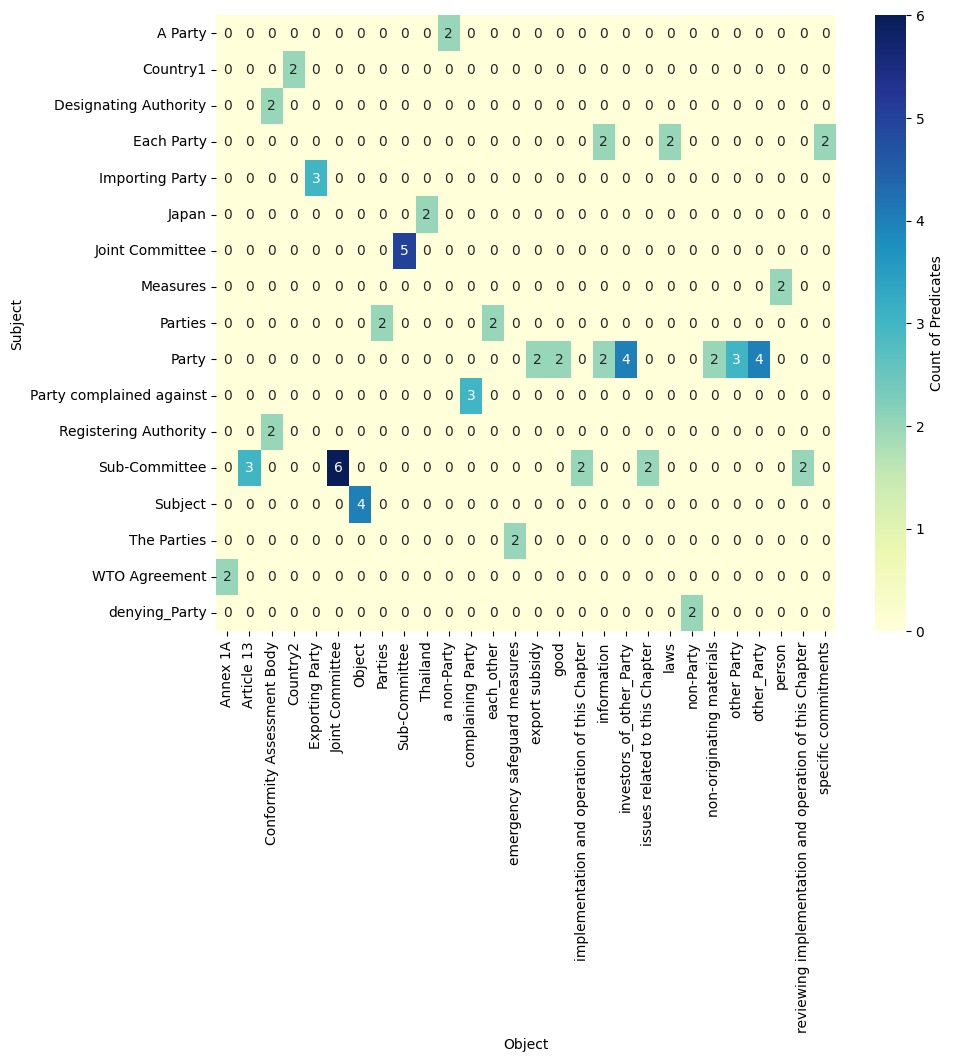}
         \caption{}
        \label{fig:heatmap1}
     \end{subfigure}
     \hfill
     \begin{subfigure}[b]{0.24\textwidth}
         \centering
         \includegraphics[width=\textwidth, height=2.5cm]{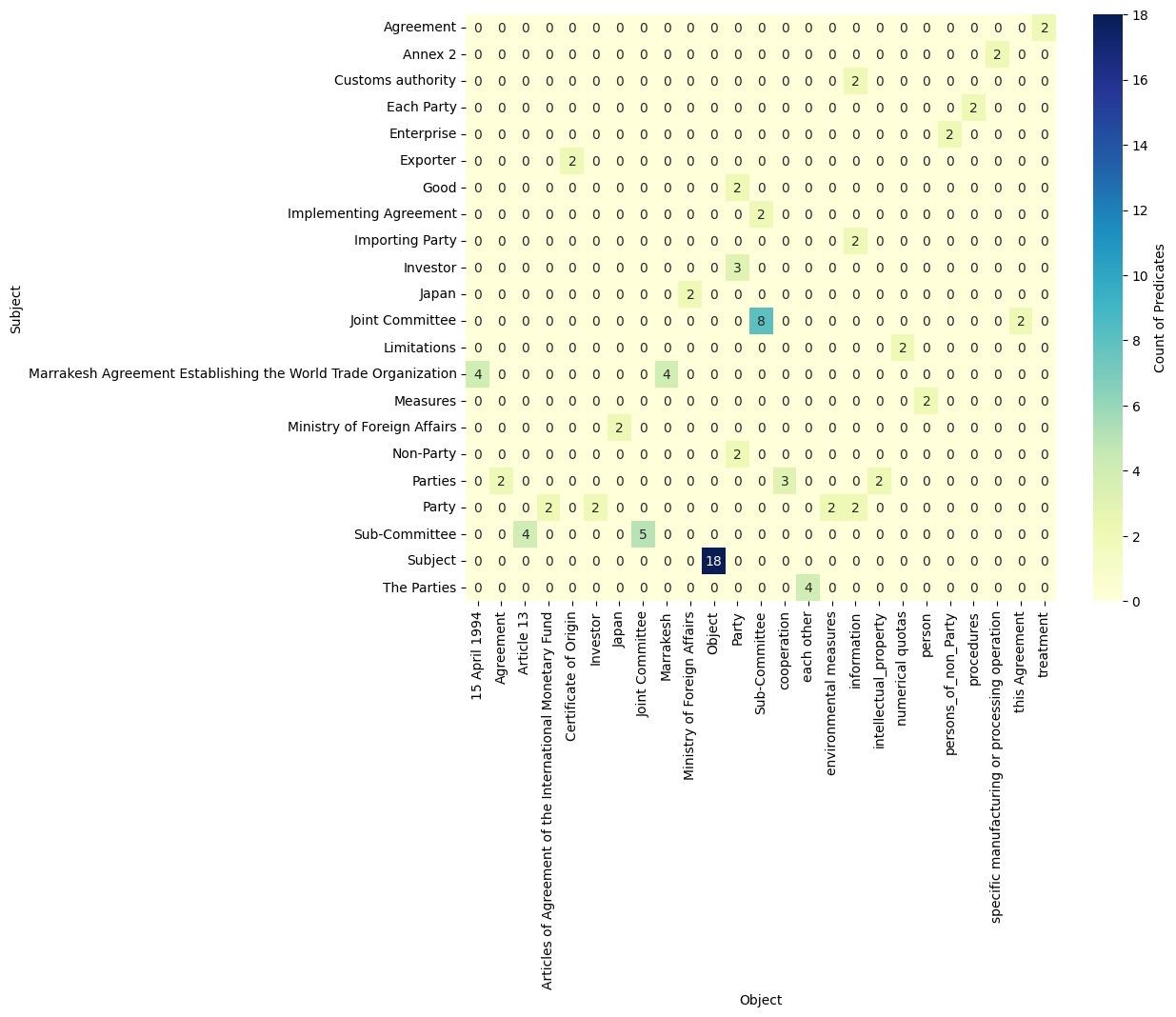}
         \caption{}
         \label{fig:heatmap2}
     \end{subfigure}
     \hfill
     \begin{subfigure}[b]{0.24\textwidth}
         \centering
         \includegraphics[width=\textwidth, height=2.5cm]{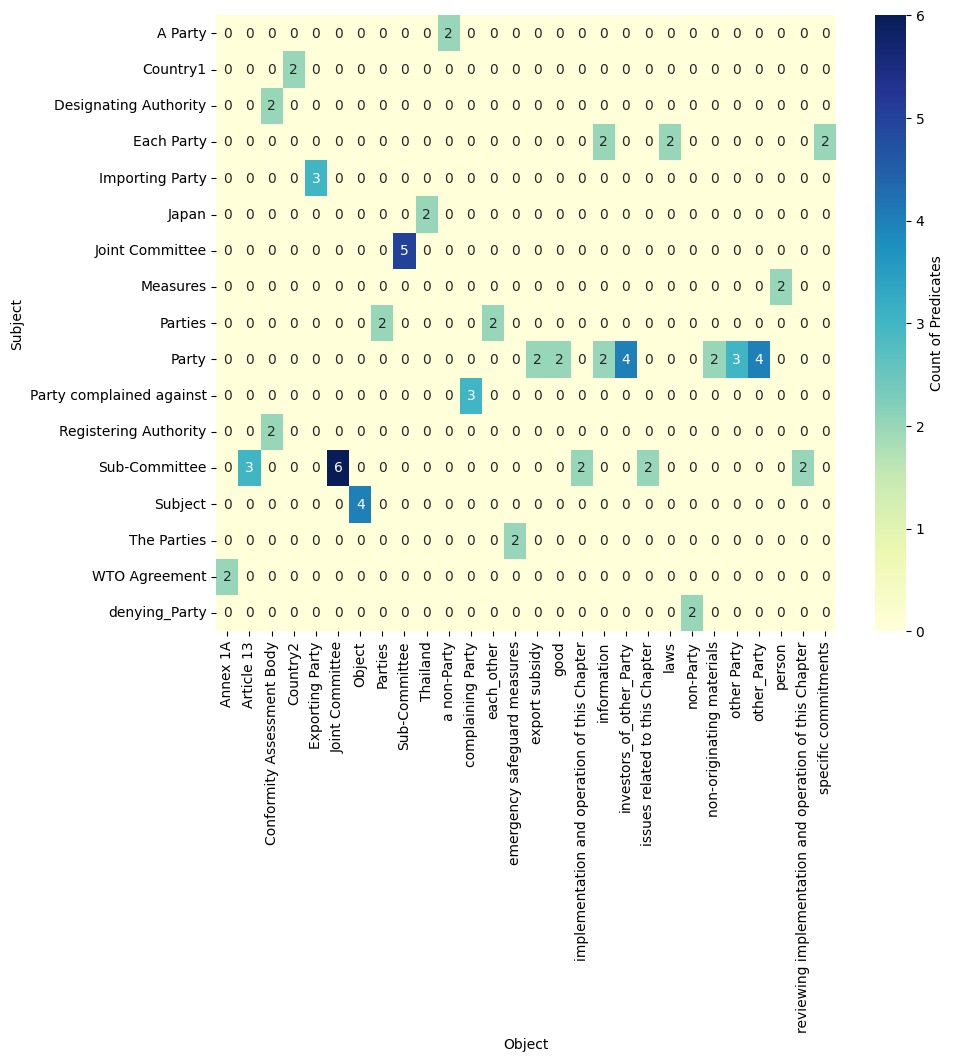}
        \caption{}
        \label{fig:heatmap3}
     \end{subfigure}
     \hfill
     \begin{subfigure}[b]{0.24\textwidth}
         \centering
         \includegraphics[width=\textwidth, height=2.5cm]{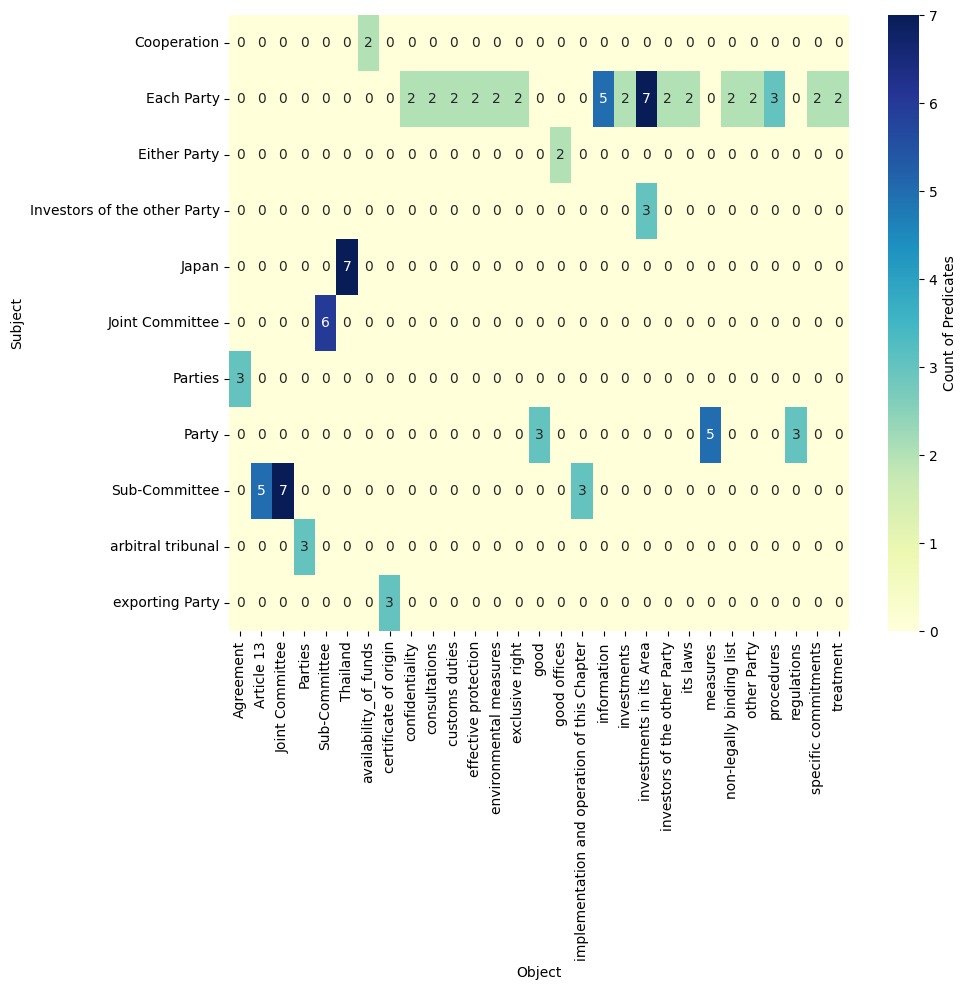}
        \caption{}
        \label{fig:heatmap4}
     \end{subfigure}
        \caption{Heatmap: (a) Zero Shot, (b) One Shot, (c) Few Shot, (d) Negative Examples}
        \label{fig:heatmap}
\end{figure}

When compared to the benchmark dataset, this distribution suggests that the Llama 3.1 model effectively captures predicates relevant to the domain but may require fine-tuning to enhance the diversity of its outputs. Next, we generated heatmaps for each prompt configuration. Figure \ref{fig:heatmap} shows the heatmaps for each of the prompt configurations. 

Lastly, we also evaluate the results through the knowledge of domain experts to qualitatively analyse the triples generated and compare the results to the benchmark triples generated by the economic trade domain experts. We do this by using the metrics defined below. 

The first metric, \textbf{Relation Validation}, shows that Llama 3.1 generated triples generally have strong contextual and semantic accuracy. Predicates like cooperate\_with, expand\_trade\_with, and invest\_in reflect the economic and diplomatic nature of the agreement, though some overly complex predicates reduce usability. The second metric, \textbf{Entity-Relation Coherence}, assesses the alignment between entities and predicates. The model typically identifies country-level actors (e.g., Japan, Thailand) correctly and maintains logical relationships, though it occasionally uses general terms like “Parties” instead of specific names. The third metric, \textbf{Triple Completeness}, evaluates whether essential information is preserved. While key themes such as trade liberalization and investment protection are captured, some higher-level strategic insights found in manual triples are missed. The fourth metric, \textbf{Semantic Correctness}, confirms that most triples are coherent, with logical subject-predicate-object structure, though some would benefit from more standardized predicate forms. The fifth, \textbf{Information Gain}, and the sixth metrics \textbf{Redundancy}, measures whether the model contributes useful new content and avoids repetition, respectively. It adds detail by breaking down complex clauses but sometimes generates redundant triples with slight predicate variations. The seventh and the eigth metrics, \textbf{Predicate Distribution} and  \textbf{Coverage} respectively, looks at how well the predicates reflect the agreement's scope and balance between countries. Most aspects like trade, cooperation, regulation—are covered, and bilateral relations are reasonably consistent, though bidirectional predicates could better represent reciprocity.

\section{Conclusion}
\label{conclusion}
In this work, we describe the methodology to extract subject-predicate-object triples from natural language texts using Llama 3.1 and experiment by iteratively improving the prompts for triple extraction and observe that including positive examples and negative examples increases the quality of the triples extracted. Through our experiments, we observed that the Llama 3.1 generated triples show strong performance in capturing detailed relationships and maintaining semantic coherence. We also observed that in order to generate qualitative results, language models require advance fine tuning and consistent feedback. A major drawback that required a lot of attention was coreference resolution and we observed that the model was insufficient when it comes to resolving conflicts with coreferences. \\

\noindent\textbf{Acknowledgement} The work has been done as the part of KONECO project, and it has received funding from the Bundesministerium f{\"u}r Bildung und Forschung (BMBF) under grant No 16DKWN095.

\bibliographystyle{splncs04}
\bibliography{references}
\end{document}